\documentclass[11pt,a4paper]{article}
\usepackage{times,latexsym}
\usepackage{url}

\usepackage{amsmath} 
\usepackage{amsfonts}
\usepackage{hhline}
\usepackage[export]{adjustbox}

\usepackage{subcaption}

\usepackage{booktabs} 
\usepackage{multirow}

\usepackage{paralist}

\usepackage[T1]{fontenc}
\usepackage{tikz}
\usetikzlibrary{arrows}
\usetikzlibrary{decorations.pathreplacing,calligraphy}
\usepackage{soul}

\usepackage{dblfloatfix}

\usepackage[acceptedWithA]{tacl2021v1}

\usepackage{xspace,mfirstuc,tabulary}

\newif\iftaclinstructions
\taclinstructionsfalse 
\iftaclinstructions
\renewcommand{\confidential}{}
\renewcommand{\anonsubtext}{(No author info supplied here, for consistency with
TACL-submission anonymization requirements)}
\newcommand{\instr}
\fi

\iftaclpubformat 

\else

\fi

\title{Generative Text Modeling and Language-Guided Visual Attention for Deciphering Palimpsests}

\author{
   Robin Hollifeldt$^\dagger$,
   Tianru Zhang$^\dagger$,
   Jessica Ström$^\dagger$ and
   Ekta Vats$^\dagger$
   \\
   \ \\
   $^\dagger$Uppsala University
   \\
   Uppsala, Sweden
   \\
   \texttt{robin.hollifeldt@it.uu.se}, \texttt{tianru.zhang@it.uu.se},
   \\
   \texttt{jessica195.js@gmail.com},
   \texttt{ekta.vats@it.uu.se}
 }

\date{}

\begin{document}
\maketitle
\begin{abstract}
Recovering obscured text in palimpsests, i.e. the manuscripts where the original writing (under-text) was erased and overwritten for reuse, is vital for preserving lost information. Due to the nature of the material, such a task can be highly challenging and time-consuming. This paper presents an approach towards automated palimpsest handwritten text modeling for under-text recovery. We investigate generative models, including Generative Adversarial Networks (GANs) and diffusion models, and design a custom denoising diffusion model that learns structured noise patterns to reconstruct the under-text. The recovered text is subsequently processed by a fine-tuned, attention-based handwritten text recognition network. Furthermore, we investigate a language-guided recognition framework that leverages textual class descriptions as informative priors, enabling the text recognition model to distinguish meaningful features from noise. This is achieved by grounding task-specific descriptions, phrased in words or sentences, through visual attention maps generated by pretrained vision-language models like CLIP (Contrastive Language-Image Pre-training). To facilitate future research and benchmarking, we release new datasets for synthetic palimpsests (currently scarce), CLIP-compatible text descriptions, and pre-trained models with code.
\end{abstract}

\section{Introduction}
\label{sec:intro}

Palimpsests are historical manuscripts that contain a visible top layer of text and one or more hidden, erased layers beneath it. The original writings were scraped-off or overwritten to reuse the writing material (parchment). Restoring palimpsests and recovering the underlying handwriting is fundamentally an inverse problem, one that aims to reverse the processes of erasure and overwriting to reveal the lost text.

While multispectral and hyperspectral imaging can enhance contrast between inks and parchment, other factors such as spectral overlap, surface irregularities, and noise often obscure the faint signals of the under-text. Because these limitations constrain the effectiveness of imaging-based techniques, they have prompted exploration of data-driven approaches, such as deep learning. However, emerging deep learning approaches face their own limitations, due to the scarcity of annotated training data, poor generalisation across different manuscripts, and the risk of generating misleading reconstructions. The field remains significantly under-explored, and to the best of our knowledge, no existing method can reliably separate over-text from under-text across diverse manuscripts and perform accurate recognition. This gap highlights the need for standardised, reproducible workflows for manuscript restoration.

Recent advances in handwritten text recognition (HTR) have leveraged natural language processing (NLP) techniques to improve recognition accuracy \cite{li2025ego, chen2024trocr}. Language models, textual priors, and semantic embeddings provide contextual information that helps disambiguate visually uncertain predictions, while vision-language frameworks enable grounding of textual descriptions in visual features. Integrating NLP with HTR transforms it into a multimodal problem, allowing models to distinguish relevant text from noise and better handle challenging scenarios such as palimpsests. 

Our work builds on this line of research and examines the extent to which natural language guidance can improve the reconstruction and recognition of under-text in palimpsests using a deep learning framework. Our key contributions: 
\begin{compactenum}[a)]
    \item We examine how generative models, including Generative Adversarial Networks (GANs) and diffusion models, can enhance under-text extraction, and introduce a custom denoising diffusion model (cDDM) tailored to this problem.
    \item We introduce a language-guided, attention-based recognition framework that leverages textual priors to guide HTR.
    \item We explore a new distillation paradigm that transfers knowledge via guided attention and integrates CLIP-based vision-language grounding to align semantic text descriptions with visual features, enabling the model to distinguish handwriting from noise.
    \item The proposed approach is model-agnostic and broadly transferable to any HTR framework that exposes intermediate activation maps and utilises a standard classification loss.
    \item To address the scarcity of training data, we create and release synthetic palimpsest datasets, CLIP-compatible text descriptions, and pre-trained models, providing resources for future research and benchmarking. 
\end{compactenum}

The source code for our implementation is publicly available at \url{https://github.com/UU-VLL/GTM-LGHTR-Palimpsest}. The dataset is available at \url{https://doi.org/10.5281/zenodo.21902987}.

\section{Related Work}
\label{sec:related}
When it comes to palimpsest restoration, traditional approaches, such as colour space transformation \cite{tonazzini2010color}, and statistical methods, such as Principal Component Analysis and Independent Component Analysis on multi-spectral imaging \cite{salerno2007digital, Easton2011}, have long been the leading techniques for text extraction and layer separation. Recently, deep learning models, including Convolutional Neural Networks \cite{perino2023deep}, generative inpainting \cite{jampour2024enhancing}, latent diffusion models \cite{jampour2025revealing}, and the increasingly popular GANs \cite{Starynska2021, madi2023scheme}, have begun to play a more important role in this field. In particular, \citet{madi2023scheme} compared several GAN models and showed that their proposed models, Palimpsest Manuscripts Reconstruction (PMR)-GAN, and Pix2pix (also a GAN model), are among the top-performing approaches in restoration. Given the difficulty of obtaining ground truth images for training palimpsest restoration models, all the aforementioned deep learning methods generated synthetic data using various techniques. Meanwhile, despite the general consensus that diffusion models may achieve better results than GANs in other fields \cite{dhariwal2021diffusion, muller2023multimodal}, there have been relatively few investigations of their application in palimpsest restoration. Latent diffusion models were investigated by \citet{jampour2025revealing}, who also introduced the Synthetic Georgian Palimpsest datasets \cite{jampour2024enhancing}. However, their evaluation was limited to only a few palimpsests, leaving it unclear how well the method generalises across diverse datasets.

Palimpsest (under-text) recognition remains an under-explored area, with the earliest studies emerging as late as 2017 \cite{starynska2017methods}, focusing primarily on character-level Optical Character Recognition (OCR). Subsequent works have also employed OCR/HTR accuracy as an indirect measure of restoration quality \cite{Starynska2021, madi2023scheme}. To the best of our knowledge, no known studies have attempted word-level under-text HTR that generalises across diverse manuscripts, and existing approaches remain confined to specific manuscript types.

Because annotated palimpsest data are scarce, current studies rely on synthetic datasets that closely imitate real manuscripts. This approach enables supervised training and quantitative evaluation of text separation models. Synthetic generation also allows controlled creation of target masks for text separation and labels for recognition.

With the emergence of the CLIP model in vision-language tasks \cite{radford2021learning}, its potential for vision-language alignment \cite{petryk2022guiding} has yet to receive substantial attention from researchers working on HTR and palimpsest recognition. Existing studies in Scene Text Recognition have demonstrated the latent OCR capabilities of CLIP, for example through the incorporation of CLIP-based cross-modal alignment plug-ins with existing detection heads \cite{yu2023turning} and the integration of CLIP encoders with multi-branch decoders \cite{zhao2024clip4str}. Other research exploring CLIP in OCR tasks has investigated innovative alignment and distillation techniques \cite{wang2023symmetrical} as well as integration with large language models \cite{zhao2024clip}.

In this work, CLIP is employed as a guidance model whose text-conditioned activation maps, obtained via gradient-weighted attention rollout \cite{chefer2021generic}, are used to supervise the Grad-CAM \cite{selvaraju2017grad} attention of the HTR model (in our case the attention-based HTR model \cite{kass2022attentionhtr}), aiming to enhance modality alignment. This work also evaluates the Pix2pix GAN model, and designs a custom diffusion model for under-text retrieval and restoration tasks. Model generalisation and reproducibility are of critical importance; therefore, we conduct a comprehensive evaluation across three datasets: Synthetic MNIST, Synthetic Saint Gall, and Synthetic Georgian Palimpsest. One of the few well-known real palimpsest datasets is the Archimedes dataset \cite{toth2008archimedes}. However, while some under-text is visible, it lacks paired samples of overlapping and clean under-text, making it unsuitable for training our generative models.

It is important to note that no sample selection is performed in this work and the evaluations are conducted across the predefined data splits in accordance with the dataset guidelines and evaluation metrics. We believe this will greatly help facilitate future research and benchmarking.

\section{Methodology}
\label{sec:main}
Our primary objective is to recognise the under-text from a synthetic palimpsest. This can be approached in two ways: directly perform HTR on the partially visible under-text, or first carry out text layer separation to reconstruct the under-text image, then perform HTR on the reconstructed image. Experiments are conducted on both methods, with the results presented later in section \ref{subsec:results}. In this section we first introduce our text separation techniques, followed by the HTR methods.

A palimpsest image \(x\) can be viewed approximately as a superposition of the over-text \(x_\text{over}\), the under-text \(x_\text{under}\) and the background writing support \(x_\text{bg}\). The text separation stage (section \ref{subsec:sep}) in our pipeline aims to leverage generative models to extract signals from the under-text as an image, and the recognition stage (section \ref{subsec:gals}) uses an HTR model to identify the text from such images with the guidance of the CLIP model. The overall architecture of the integrated pipeline is shown in Figure \ref{g-ahtr_architecture}. The following subsections describe the text separation and HTR models.

\subsection{Palimpsest Text Separation}
\label{subsec:sep}

The text separation process can be expressed as training a model that takes the palimpsest image \(x\) as input, and reconstructs the under-text \(x_\text{under}\). In the subsequent section we will denote the ground truth under-text \(x_\text{under}\) as the training target $y$, and introduce the two text separation models implemented in this work.

\subsubsection{Pix2pix}
Pix2pix is a conditional GAN with skip connections, demonstrating state-of-the-art performance in text separation for palimpsests on synthetic data \citep{madi2023scheme}. The architecture is well-suited for image-to-image translation, enabling the separation of under-text, over-text, and background. Traditional methods often fail in overlapping regions \citep{Starynska2023}, whereas Pix2pix preserves structural integrity through its U-Net generator and adversarial training, yielding higher-quality under-text reconstruction. The adversarial loss further enforces realism in the generated outputs \citep{isola2017pix2pix}, allowing conditional GANs to model inter-pixel dependencies absent in standard GANs.

Conditional GANs such as Pix2pix learn to map an input image \(x\) to a target image \(\hat{y}\) through two competing networks: a generator \(G\) and a discriminator \(D\). \(G\) aims to produce \(\hat{y}\) indistinguishable from the target real image \(y\), while \(D\) learns to differentiate between them. Their adversarial interplay leads both networks toward optimality.

Pix2pix employs a U-Net generator, an encoder-decoder architecture that first downsamples the input feature \(x\) and then reconstructs through upsampling. The forward pass can be expressed as
\begin{equation}
\scalebox{0.9}{$
x \xrightarrow{G_{Enc_1}} f_1 \xrightarrow{\cdots} \xrightarrow{G_{Enc_n}} f_n \xrightarrow{G_{Dec_1}} f_{n+1} \xrightarrow{\cdots} \xrightarrow{G_{Dec_n}} \hat{y}$}
\label{eq:u_Net}
\end{equation}

\noindent where \(f_n\) denotes the bottleneck feature representation, $G_{Enc_i}$ and $G_{Dec_j}$ represent the $i$th encoder and the $j$th decoder in the U-Net. 

The discriminator \(D\) is implemented as a PatchGAN, dividing each image into local patches and predicting a real/fake probability per patch:
\begin{equation}
\begin{aligned}
(x, \hat{y}) &\xrightarrow{D_{Conv_1}} f_1 \xrightarrow{\cdots} \xrightarrow{D_{Conv_n}} \{p_i^{\text{fake}}\}_{i=1}^N,\\
(x, y) &\xrightarrow{D_{Conv_1}} f_1 \xrightarrow{\cdots} \xrightarrow{D_{Conv_n}} \{p_i^{\text{real}}\}_{i=1}^N,
\label{eq:patchGAN}
\end{aligned}
\end{equation}
where \(Conv_{1...n}\) stands for convolutional layers whose final output is a map of real-valued patch scores \(p_i\), and \(N\) stands for the number of patches in each image. The least-squares objectives below are applied to the patch scores directly, i.e.\ the loss is the mean over the $N$ patches of the per-patch squared errors, $\frac{1}{N}\sum_{i=1}^{N}(p_i - c)^2$ with target $c \in \{0, 1\}$.

To stabilise training, Pix2pix applies the Least Squares GAN (LSGAN) loss \citep{mao2017least}, which replaces the binary cross-entropy (CE) loss \citep{goodfellow2014generative} with a least-square objective for stronger gradient feedback.

The generator loss combines adversarial and reconstruction terms:
\begin{equation}
\begin{aligned}
\mathcal{L}_G = \mathbb{E}_{x}\left[(D(x, G(x)) - 1)^2\right] 
\\
+ \lambda\, \mathbb{E}_{x,y}\left[\|y - G(x)\|_1\right],
\label{eq:Gloss}
\end{aligned}
\end{equation}
where \(\lambda=10\) balances the realism and pixel-level accuracy; note that this is a smaller reconstruction weight than the \(\lambda=100\) used by \citet{isola2017pix2pix}.
The discriminator loss is defined as
\begin{equation}
\begin{aligned}
\mathcal{L}_D =\;& \tfrac{1}{2}\mathbb{E}_{x,y}\!\left[(D(x,y) - 1)^2\right] \\
&\quad + \tfrac{1}{2}\mathbb{E}_{x}\!\left[(D(x, G(x)))^2\right].
\end{aligned}
\label{eq:Dloss}
\end{equation}
Through this loss, \(D\) learns to output values close to 1 for real samples and close to 0 for generated ones, ensuring effective adversarial training.

\begin{figure*}[!t]
    \centering
    \begin{tikzpicture}[
        very thick,
        boxnode/.style={draw, rounded corners, rectangle, inner sep=6pt}
        ]
      \draw[boxnode]
        (-3.5,-2.8) node[] (palimpsest) {\includegraphics[width=.1\textwidth]{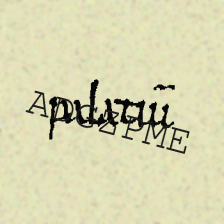}}
        (-3.5,-4.0)  node[] (palimpsest_image) {Palimpsest image}
        (-3.5,0)  node[draw, fill=yellow!40, text width=1.6cm, align=center] (sep_model) {Text separation model}
        (-0.5,0) node[] (image_ex) {\includegraphics[width=.1\textwidth]{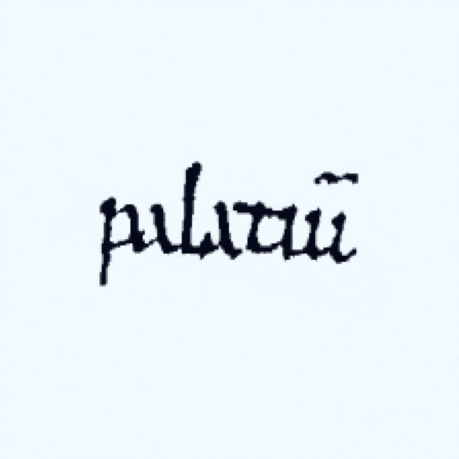}}
        (-0.5,-1.2)  node[] (image) {Under-text image}
        (-0.5,-2) node[text width=4cm, align=center] (text_ex) {\texttt{\scriptsize A photo of a hand- written word palatiu}}
        (-0.5,-2.8) node[] (text) {Input guiding text}
        (3,0)  node[draw, fill=red!20] (model) {HTR model}
        (3,-2) node[draw, fill=blue!15] (CLIP) {CLIP}
        (7,1) node[] (gt) {text label: \texttt{palatiu}}
        (7,0) node[] (ot) {text pred: \texttt{palatw}}
        (7,-1) node[] (om) {HTR activation map \(\mathcal{A}^\text{HTR}\)} 
        (7,-2) node[] (gm) {CLIP guidance map \(\mathcal{A}^\text{gt}\)} 
        (7,-3.2) node[] (map_ex) {\includegraphics[width=.1\textwidth]{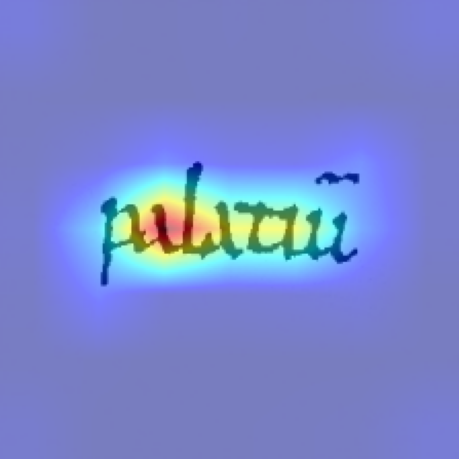}}
        (10,0.5) node[] (loss1) {CE loss}
        (10,-0.5) node[] (plus) {+}
        (10.2,-1.3) node[] (loss2) {activation}
        (10.2,-1.7) node[] (loss3) {loss};
    
      \draw [->] (palimpsest) to (sep_model);
      \draw [->] (sep_model) to (image_ex);
      \draw [dashed] (palimpsest) to [out=50,in=180] (image_ex);

      \draw [->] (image_ex) to (model);
      \draw [->] (image_ex) to [out=0,in=180] (CLIP);
      \draw [->] (text_ex) to (CLIP);
    
      \draw [->] (model) to (ot);
      \draw [->] (model) to [out=0,in=180] (om);
      \draw [->] (CLIP) to (gm);

      \coordinate (m1) at (6,1.7);
      \draw[rounded corners=10pt,->] (loss1) |- (m1) -| (model);
    
      \draw [decorate,
        decoration = {calligraphic brace}] (9.0,1) -- (9.0,0);
        \draw [decorate,
        decoration = {calligraphic brace}] (9.2,-1) -- (9.2,-2);
      
    \end{tikzpicture}

    \vspace{-0.1in}
    \caption{Architecture of our proposed framework, consisting of two main stages: \textbf{(i) Under-text separation}, where a trained generative model is used to recover the target under-text from the palimpsest. \textbf{(ii) Language-guided HTR}, where an HTR model is trained using a standard CE loss between the predicted text and the ground truth. To enhance recognition, we introduce a language-guided supervision strategy, which aligns the HTR model's Grad-CAM visual feature maps with text-image alignments derived from CLIP via an activation loss. The example palimpsest image originates from the Synthetic Saint Gall dataset, and the example under-text image is a reconstructed image using cDDM2.}
    \label{g-ahtr_architecture}
\end{figure*}

\subsubsection{cDDM}
\label{subsubsec:cDDM}
Historical palimpsests are highly diverse and contain degraded script where the unwanted components (e.g., over-text and complex background noises) are highly structured and do not follow a Gaussian distribution. Therefore, standard Denoising Diffusion Probabilistic Models (DDPM) \cite{ho2020denoising}, which rely on stochastic Gaussian noise mapping, may discard relevant information when applied to palimpsests.
In this work, we propose a custom conditional Denoising Diffusion Model (cDDM). Instead of using randomly sampled noise, cDDM adapts the diffusion architecture to invert a fixed, structured residual and operates as a deterministic diffusion process. This framing aligns with the generalised diffusion literature (e.g., Cold Diffusion \cite{bansal2023colddiffusion}), which demonstrates that the diffusion framework can successfully invert arbitrary, non-probabilistic image degradations without relying on Gaussian assumptions.

In traditional DDPMs, the Gaussian noise $\epsilon \sim \mathcal{N}(0, I)$ is added in each time step via
\begin{equation}
x_t = \alpha_t x_{t-1} + \sqrt{\beta_t} \epsilon,
\label{eq:5}
\end{equation}
where $\alpha_t$ controls the signal retention, $\sqrt{\beta_t}$ controls the noise scale, and $t=1{:}T$ indexes the diffusion steps (with $x_0$ denoting the clean image).
To address the structured, non-Gaussian artifacts in our case, we replace stochastic sampling with a deterministic residual between the raw palimpsest input $x$ and the target under-text $y$.
\begin{equation}
x_t = \alpha_t x + (1-\alpha_t)\, y = y + \alpha_t \epsilon, \quad \epsilon = x - y,
\label{eq:6}
\end{equation}
i.e. a convex combination of the palimpsest $x$ and the target $y$, which keeps the target $y$ and the full structured residual $\epsilon$ present in every latent $x_t$. Because $\beta_t$ is clipped to $[10^{-4},\, 2\times 10^{-2}]$ (see Eq. \ref{eq:beta}), the coefficient $\alpha_t=\sqrt{1-\beta_t}\in[0.990,\,1.000]$ varies only mildly with $t$: every $x_t$ remains close to the palimpsest image $x$, and the model is trained, at all noise levels, to regress the complete structured residual $\epsilon = x - y$ from an input that is approximately $x$. The time step therefore acts primarily as an auxiliary conditioning signal (via the time embedding) rather than as an interpolation between $y$ and $x$; this differs from standard DDPMs, in which $x_T$ is (almost) pure noise. In particular, $x_t$ never degenerates to pure noise, so the structured, non-Gaussian residual is available to the network at every step. Input images are normalised to $[-1,1]$ to capture fine differences, e.g., between two black texts.

The residual is defined in the raw pixel space, where \(y\) denotes the clean target under-text image that the model is trained to recover from the palimpsest images \(x\). At each step, the noise $\epsilon$ is given by \(x - y\), which represents the unwanted components in \(x\), such as the background and over-text that are treated as noise.

The rate at which the noise scale varies is controlled via a cosine noise scheduler \cite{nichol2021improved}. The cosine scheduler is defined as
\begin{equation}
\bar{\alpha}_t = \frac{f(t)}{f(0)}, \quad f(t) = \cos^2\left( \frac{t/T + s}{1+s} \cdot \frac{\pi}{2} \right),
\end{equation}
where $s = 0.008$ is a small offset that prevents $\beta_t$ from becoming too small near $t=0$ \cite{nichol2021improved}, and per-step noise scales
\begin{equation}
\begin{aligned}
\beta_t = \operatorname{clip}\!\left(1 - \frac{\bar{\alpha}_{t}}{\bar{\alpha}_{t-1}},\; 10^{-4},\; 2\times 10^{-2}\right),
\\
\quad \alpha_t = \sqrt{1-\beta_t},
\end{aligned}
\label{eq:beta}
\end{equation}
where the clipping bounds, adopted from the linear-schedule DDPM convention, cap the per-step noise change; for the small $T$ used in cDDM1 they are active at almost every step, so $\alpha_t$ is nearly constant ($\approx 0.99$).

Subsequent to the forward process that progressively adds noise, the reverse process is designed to remove noise and reconstruct the original data, through training a deep learning model to approximate the noise and ultimately learn to remove it from a latent variable \(x_t\). To do so, a U-Net is employed to train on $(x_t, \textit{Embedding}(t))$, where time $t$ is embedded into a vector to condition the model on the time state. The network flow of the model can be expressed as follows:
\begin{equation}
\begin{aligned}
(x_t, \textit{Embedding}(t)) \xrightarrow{E_1} f_1 \xrightarrow{\cdots}  \xrightarrow{E_n} f_n \\
\xrightarrow{D_1} f_{n+1} \xrightarrow{\cdots}  \xrightarrow{D_n} \hat{\epsilon},
\label{eq:DDPMforward}
\end{aligned}
\end{equation}
where $E_{1...n}$ and $D_{1...n}$ are the encoders and decoders of the U-Net, and $\hat{\epsilon}$ is the predicted noise. The model is then trained on minimising the $L_1$ loss:
\begin{equation}
    \mathcal{L} = \mathbb{E}_{x, y, t} \left[ \left\| \hat{\epsilon} - \epsilon \right\|_1 \right].
\label{eq:l1loss}
\end{equation}
Note that, unlike in standard DDPMs, $\epsilon = x - y$ is fully determined by the training pair $(x, y)$, so the expectation is taken over the data and the time step only. At each training iteration, \(t\) is randomly sampled from a discrete uniform distribution: $t \sim \text{Uniform}\{1, \ldots, T\}$ (the implementation uses the equivalent 0-based indices $t-1 \in \{0, \ldots, T-1\}$).

In the inference stage, the trained U-Net $f(\cdot, \cdot)$ is used to recover previously unseen under-texts directly from the palimpsest image $x$, which is treated as the maximally noisy variable $x_T$, without access to $y$. In this work we design two denoising methods, cDDM1 and cDDM2. In cDDM1, starting from $x_T = x$, the model re-estimates the noise from the current iterate at every step and removes it deterministically according to:
\begin{equation}
\begin{aligned}
\hat{\epsilon}_t &\leftarrow f(x_{t}, t), \\
x_{t-1} &\leftarrow \frac{x_t - \sqrt{\beta_t} \cdot \hat{\epsilon}_t}{\alpha_t} \quad (t = T, \ldots, 2), \\
\hat{y} &\leftarrow x_1 - \hat{\epsilon}_1,
\end{aligned}
\label{eq:inf1}
\end{equation}
i.e. the deterministic inversion of the standard per-step update (Eq.~\ref{eq:5}) with $\sqrt{\beta_t} = \sqrt{1-\alpha_t^2}$, applied with the noise re-estimated from the current iterate, and with the predicted noise removed directly in the final step. Note that this reverse recursion is not the exact algebraic inverse of the forward construction in Eq.~\ref{eq:6}; because $\hat{\epsilon}$ is re-predicted from $x_t$ at every step, the procedure acts as an iterative residual-removal scheme whose per-step magnitude follows the forward cosine schedule.

cDDM2 performs a single-step reverse process, predicting and directly removing noise in $x_T$. To improve denoising, a mask sets regions with high predicted noise to white, heuristically targeting over-text that is otherwise difficult to remove. The process is:
\begin{equation}
\begin{aligned}
    \hat{\epsilon}_T &\leftarrow f(x_T, T), \\
    \hat{y} &= x_T - \hat{\epsilon}_T, \\
    m_{i,j} &= \mathbb{I}\!\left[\operatorname{mean}_c\big( |\hat{\epsilon}_T(i,j,c)| \big) > \delta\right], \\
    \hat{y}'_{i,j} &= m_{i,j} + (1 - m_{i,j}) \cdot \hat{y}_{i,j},
\label{eq:inf2}
\end{aligned}
\end{equation}
where $\mathbb{I}[\cdot]$ is the indicator function, the predicted-noise magnitude is averaged over colour channels $c$, $m_{i,j}$ masks pixels whose value exceeds the threshold $\delta$ (setting them to white), and $\hat{y}'$ is the final denoised image. The threshold is set per dataset to $\delta = 0.5$ (Synthetic MNIST), $0.8$ (Synthetic Saint Gall) and $0.9$ (Synthetic Georgian), chosen on the training data. Unlike iterative cDDM1, noise scaling from the forward schedule is not applied in this single-step denoising approach.

To the best of our knowledge, this is the first diffusion-based approach applied directly in the raw pixel space for palimpsest text separation. While \citet{jampour2025revealing} focuses on computational efficiency, our cDDM is specifically designed to preserve high-frequency handwriting features, allowing us to benefit from diffusion's noise removal capabilities without sacrificing fine detail.

\subsection{Language-Guided Text Recognition}
\label{subsec:gals}
The palimpsest HTR learning problem can be formulated as follows. Given a palimpsest dataset for an HTR task $\mathcal{T}$, let each image contain some region of pixels that is irrelevant to the task (e.g. over-text, damages, degradations) and a region of pixels that is relevant (e.g. under-text). The goal is to recognise the under-text from certain regions in the under-text image (either the raw image or the reconstructed image from the text separation). 

In this work, we propose language-guided HTR that leverages a vision-language model to guide a conventional HTR system by constraining it to focus on a more semantically grounded attention map. The architecture is shown in the right half of Figure \ref{g-ahtr_architecture}, starting from the under-text image that has been recovered from a given input palimpsest image using the text separation model discussed in the previous section (Pix2pix or cDDM).

For every image \(y\) (or retrieved image \(\hat{y}\)) and ground truth text label \(y_\text{label}\), we create a natural language guiding text \(\tau\) that captures the key semantic concept relevant to $\mathcal{T}$. This guiding text is derived from the dataset's ground truth label. To ensure compatibility with a pretrained vision-language model like CLIP, \(\tau\) consists of descriptive phrases formatted similarly to CLIP's default notation. For example, we prepend a structured prompt such as ``A photo of a handwritten word $label$'', where the word $label$ is extracted from the dataset's ground truth text label \(y_\text{label}\).

For an HTR model $f_\theta$ with parameters $\theta$, we aim to learn an optimal HTR model that outputs a text prediction and an activation map for a given under-text image as input. To learn an optimal $f_\theta$, we can estimate a probability map over the latent space, where each entry corresponds to the probability that pixel is relevant to $\tau$. In this work, we build on the AttentionHTR model \cite{kass2022attentionhtr} (AHTR in short). In principle, any HTR model that can generate intermediate activation maps and support a standard classification loss could be used.

Next, we introduce language-guided supervision by generating ground truth activation maps from CLIP. We refer to the resulting model as G-AHTR (Guided-AttentionHTR). CLIP is a vision-language model that takes both an image and a text input and produces a similarity score in a shared embedding space. Although this score is global, gradient-based attribution allows us to propagate it through CLIP's visual encoder to obtain pixel-level relevance maps \cite{petryk2022guiding}, enabling us to exploit CLIP's text-image alignment without modifying the base HTR architecture. Since our CLIP visual encoder is a ViT-B/32 transformer, we compute these maps with gradient-weighted attention rollout \cite{chefer2021generic}, accumulating the gradient-weighted attention of all transformer blocks into a relevancy map over the visual tokens.

Before extracting activation maps, we fine-tune the full CLIP ViT-B/32 model on the training split of the respective (reconstructed) dataset with the standard symmetric image--text contrastive loss (temperature $0.07$), using AdamW with learning rate $10^{-6}$ and weight decay $10^{-4}$ for 20 epochs; the HTR-sized grayscale images are replicated to three channels and bilinearly resized to $224\times224$.

Given the under-text image $y$ and guiding text $\tau$, we feed both into CLIP (input resized to $224\times224$) and obtain a text-conditioned relevance map over the visual tokens, which for the ViT-B/32 encoder natively lies on a $7\times7$ grid. This map is bilinearly upsampled to the image resolution, undergoes the same crop-and-pad preprocessing as the input image (Section~\ref{subsec:datasets}), and is bilinearly resampled to the HTR input grid, yielding \[\mathcal{A}^\text{CLIP}(y, \tau): \{1,\ldots,32\} \times \{1,\ldots,100\} \to [0, 1].\] This map serves as the \textit{ground truth activation map}, denoted $\mathcal{A}^\text{gt} \equiv \mathcal{A}^\text{CLIP}$ in the following. Importantly, we do not use CLIP's prediction itself, only its activation map to provide attention guidance. In parallel, we extract an activation map from the HTR model with Grad-CAM \cite{selvaraju2017grad} at the last convolutional block of its feature extractor, using the summed logits of the ground truth character sequence as the class score (native resolution $4\times26$), bilinearly interpolated and min-max normalised to the same grid: \[\mathcal{A}^\text{HTR}(y): \{1,\ldots,32\} \times \{1,\ldots,100\} \to [0, 1].\] The L1 activation loss is computed on this common $32\times100$ grid.
Before entering the loss, the guidance map is affinely rescaled to the input value range, $\tilde{\mathcal{A}}^\text{gt} = 2\mathcal{A}^\text{gt} - 1 \in [-1, 1]$, while the predicted map $\mathcal{A}^\text{HTR}$ remains in $[0, 1]$. This asymmetric scaling serves two purposes: it amplifies the guidance gradients relative to matched-range maps, and, since the predicted map cannot attain the negative target values assigned to low-relevance regions, it persistently drives the predicted attention toward zero outside the text region, acting as a sharpening prior on the guidance.

The HTR model $f_\theta$ is then trained on minimising two objectives: (i) the original CE loss between the predicted text and the ground truth transcription, and (ii) a complementary activation loss between the HTR and CLIP activation maps. The \textit{activation loss} refers to the difference between the two activation maps, in the form of L1 norm. An unweighted sum of the two constitutes the final training objective:
\begin{align*}
    \mathcal{L}(\hat{y}_\text{label}, y_\text{label}, x, \tau)
    = &\mathcal{L}^\text{HTR}(\hat{y}_\text{label}, y_\text{label}) \\
    \quad + &\mathcal{L}^\text{activation}\Big(
        \tilde{\mathcal{A}}_{i,j}^\text{gt}(y, \tau), \\
    &\qquad
        \hspace{23pt}\mathcal{A}_{i,j}^\text{HTR}(y)
    \Big)
\end{align*}
where \(\hat{y}_\text{label}\) is the predicted transcription, \(y_\text{label}\) is the ground truth text, \(x\) is the input image, and \(\tau\) is the guiding text. Here, \(\tilde{\mathcal{A}}_{i,j}^\text{gt}(y, \tau)\) denotes the rescaled ground truth activation map extracted from CLIP, and \(\mathcal{A}_{i,j}^\text{HTR}(y)\) is the activation map from the HTR model.

\section{Experimental Results}
\label{sec:exp}
The proposed methods are empirically evaluated on three synthetic datasets using standard evaluation metrics. This section is organised as follows. We begin by presenting the synthetic datasets and detailing the process to create them. Next, we describe the evaluation metrics used. Finally, we present the results for both text separation and recognition across three scenarios: (a) recognition from the original datasets (MNIST and Saint Gall) without overlapping text, serving as a baseline; (b) direct recognition of the partially visible under-text from palimpsest images; and (c) recognition following text layer separation, where the under-text is first reconstructed.

\subsection{Datasets}
\label{subsec:datasets}

\subsubsection{Synthetic MNIST}
We replicate the Synthetic MNIST dataset used by \citet{Starynska2021} with both diversity and manageable file size in mind. The original MNIST digits \cite{deng2012mnist} serve as target under-texts. To emulate palimpsest data, where backgrounds are typically lighter than text, each image \(I(x, y)\) is inverted, $I(x, y) = 255 - I(x, y)$, where \(x, y\) denote pixel coordinates and intensities are in $[0, 255]$.

The target images are randomly rotated by \(\theta \sim \text{Uniform}(-20^\circ, 20^\circ)\) using bi-cubic resampling:
\begin{equation}
\begin{aligned}
    I'(x', y') &= I(R_\theta^{-1}(x', y')), \\
    \begin{bmatrix} x' \\ y' \end{bmatrix} &= 
    \begin{bmatrix} \cos\theta & -\sin\theta \\ \sin\theta & \cos\theta \end{bmatrix}
    \begin{bmatrix} x - x_c \\ y - y_c \end{bmatrix} +
    \begin{bmatrix} x_c \\ y_c \end{bmatrix}.
\label{eq:thet}
\end{aligned}
\end{equation}

The target text is assigned a random low-intensity uniform colour \(i \sim \text{Uniform}(0, 50)\) in RGB channels:
\begin{equation}
I_{C_{\text{RGB}}}(x, y) = [i, i, i],
\label{eq:color}
\end{equation}
and a fourth transparency channel preserves the original MNIST pixel intensities:
\begin{equation}
I_{C_\alpha}(x, y) = 1 - \frac{I(x, y)}{255}.
\label{eq:alfa}
\end{equation}

Backgrounds with parchment-like colours are generated with channel-wise uniform noise:
\begin{align}
I_{C_R}(x, y) &\sim \text{Uniform}(200, 250), \notag \\
I_{C_G}(x, y) &\sim \text{Uniform}(220, 230), \notag \\
I_{C_B}(x, y) &\sim \text{Uniform}(150, 190).
\label{eq:parch}
\end{align}

Under-texts are extracted using a threshold \(I(x, y) < 200\) and centred on the background. Over-texts consist of random Latin characters (A–Z) with random font, size (20-25), colour (Eq.~\ref{eq:color}), and rotation (Eq.~\ref{eq:thet}). Finally, additional noise \(n \sim \text{Uniform}(1, 10)\) is applied:
\begin{equation}
\begin{aligned}
N_C(x, y) &\sim \text{Uniform}(-n, n), \\
I_C'(x, y) &= I_C(x, y) + N_C(x, y).
\label{eq:nois}
\end{aligned}
\end{equation}

The dataset contains 60 000 training and 10 000 test samples, following MNIST's original split.

\subsubsection{Synthetic Saint Gall}

The Saint Gall dataset \cite{fischer2011transcription} originates from a 9th century Latin manuscript and comprises 60 pages of Carolingian minuscule script by a single scribe. It contains 1 410 text lines and 11 597 words. Images were binarised, normalised, and paired with sentence-level labels. Its historical significance, complex script, and ink-on-parchment characteristics make it a valuable benchmark for HTR models.

A synthetic version was generated to introduce real historical text and extend processing to larger images with longer words. Under-text was extracted at the word level using provided bounding boxes, separated from the background with a threshold of 100, and placed on a transparent 350\(\times\)350 canvas. Words exceeding 350 pixels in width were skipped. Under-text colour and rotation followed Eq.~\ref{eq:color} and Eq.~\ref{eq:thet}.

Backgrounds were generated similarly to the synthetic MNIST dataset with 350\(\times\)350 pixel canvases, using Eq.~\ref{eq:parch} for color, uniform noise \(n \sim \text{Uniform}(1,50)\) (Eq.~\ref{eq:nois}), salt-and-pepper noise \(\sim \text{Uniform}(0,0.04)\), and Gaussian noise with 50\% probability, \(\sigma \sim \text{Uniform}(1,3)\). Under-text was centred and images resized to 224\(\times\)224 to reduce computation while preserving quality. Over-texts consisted of random Latin strings with random fonts, size 30-40, overlaid on under-text. 

Prior to recognition, all images are dynamically cropped to ensure a text-focused region and prevent geometric distortion. Since our HTR model has a fixed $100\times32$ input resolution, directly resizing the original $224\times224$ images severely compressed the text due to excess vertical white space. To address this, we use Otsu's thresholding to detect the ink pixels and establish a bounding box. This box is then padded with background pixels to exactly match the target $100\times32$ aspect ratio before final resizing, ensuring the morphological features of the handwriting remain strictly undistorted.\\
The final synthetic dataset comprises 6 514 training and 1 050 test images; for HTR training, the training images are further split 90/10 into training and validation subsets, with the validation subset used for model selection.

\begin{figure}[!t]
  \centering{
    \includegraphics[width=0.24\linewidth]{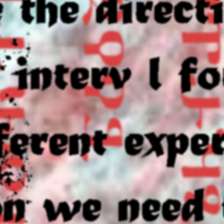}
    \includegraphics[width=0.24\linewidth]{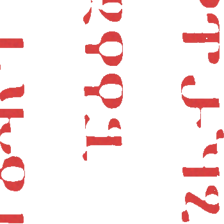}
    \includegraphics[width=0.24\linewidth]{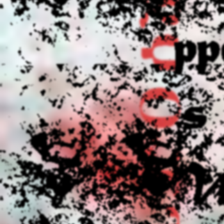}
    \includegraphics[width=0.24\linewidth]{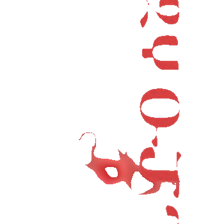}
    }
  \vspace{-0.2in}
  \caption{Combined and under-text patches from the processed synthetic Georgian dataset.}
  \label{fig:synGeorg}
\end{figure}

\subsubsection{Synthetic Georgian Palimpsest dataset}
The synthetic Georgian dataset \cite{jampour2024enhancing}, used in this work, simulates multi-spectral samples from the real Georgian palimpsest \cite{SinaiedLibrary}. Under-texts consist of 7th-8th century Caucasian Albanian text from Sinai Georgian NF 13 and 55, while backgrounds were cropped from text-free areas of the real palimpsest and prepared via stitching and cleaning. Random over-text was added, but only under-texts are considered in this work.

Targets with white backgrounds were generated by extracting under-texts using a channel-wise threshold, as in Eq.~\ref{eq:redtresh}:
\begin{equation}
\scalebox{0.8}{$
\begin{aligned}
    M_C(x, y) &= 
    \begin{cases}
    1, & \text{if } \begin{bmatrix}205 \\ 0 \\ 0\end{bmatrix} \leq I_C(x, y) \leq \begin{bmatrix}255 \\ 100 \\ 100\end{bmatrix},\\
    0, & \text{otherwise},
    \end{cases} \\
    I_C(x, y) &= M_C(x, y) \cdot I_C^{orig}(x, y) + (1 - M_C(x, y)) \cdot 255.
\end{aligned}
$}
\label{eq:redtresh}
\end{equation}

\vspace{0.05in}

Original images were 2100\(\times\)2800 pixels, too large for training. Both combined and under-text images were split into 48 patches of 350\(\times\)350, then resized to 224\(\times\)224 pixels to preserve readability. Full-image resizing rendered text unreadable. Examples are shown in Figure~\ref{fig:synGeorg}, with rare cases of slight over-extraction due to red intensity in the background.

The final synthetic Georgian dataset used in this work has 43 200 patches of training data and 4 800 patches of testing data.

\subsection{Evaluation Metrics}
\label{subsec:metric}
\paragraph{Text separation:} To assess image quality for the text separation models Pix2pix and cDDM, we use standard metrics: Structural Similarity Index Measure (SSIM), Peak Signal-to-Noise Ratio (PSNR), and the Fr\'echet Inception Distance (FID). 

\textbf{SSIM} \citep{1284395} quantifies local similarity between generated images \(\hat{y}\) and target images \(y\) based on luminance, contrast, and structure:
\begin{equation*}
 \text{SSIM}(\hat{y}, y) = \frac{(2\mu_{\hat{y}} \mu_y + C_1)(2\sigma_{\hat{y}y} + C_2)}{(\mu_{\hat{y}}^2 + \mu_y^2 + C_1)(\sigma_{\hat{y}}^2 + \sigma_y^2 + C_2)}.
\end{equation*}
Here, \(\mu_{\hat{y}}, \mu_y\) denote image means (luminance), \(\sigma_{\hat{y}}, \sigma_y\) their standard deviations (contrast), \(\sigma_{\hat{y}y}\) the covariance (structure), and \(C_1, C_2\) are small constants that stabilise the division. SSIM takes values in \([-1, 1]\), where \(1\) indicates perfect similarity.

\textbf{PSNR} \citep{wang_bovik_2009} measures reconstruction quality using the mean squared error (MSE):
\begin{align*}
   \textit{MSE}(\hat{y}, y) &= \frac{1}{nm} \sum_{i=1}^{n} \sum_{j=1}^{m} [y(i,j) - \hat{y}(i,j)]^2, \\
   \text{PSNR}(\hat{y}, y) &= 10 \log_{10} \left( \frac{L^2}{\textit{MSE}(\hat{y}, y)} \right).
\label{eq:PSNR}
\end{align*}
where \(L\) is the maximum pixel value, and \(n,m\) is the image dimensions. Higher PSNR indicates better fidelity; values above \(40\) suggest high-quality reconstruction.

\textbf{FID} \citep{heusel2018ganstrainedtimescaleupdate} measures similarity between feature distributions extracted from the 2048-d pool3 features of a pretrained Inception-v3 network (pytorch-fid), assuming Gaussian statistics:
\begin{align*}
    \text{FID}(X_y, X_{\hat{y}}) =& \|\mu_y - \mu_{\hat{y}}\|_2^2\\
    +& \operatorname{Tr}\!\left( \Sigma_y + \Sigma_{\hat{y}} - 2(\Sigma_y \Sigma_{\hat{y}})^{1/2} \right).
\end{align*}
Lower FID values indicate that generated images are closer to the real image distribution.

\paragraph{Text recognition:} Recognition performance for AHTR and G-AHTR is measured using \textbf{Word Accuracy} (WAcc), computed as: 
\begin{align*}
\text{WAcc (\%)} = 100 - \text{WER (\%)},
\end{align*}
where, since recognition is performed at the isolated-word level, the Word Error Rate (WER) is the percentage of test words that are not recognised exactly. We also compute Normalised Edit Distance (NormED) and Character Error Rate (CER) to further analyse model performance. \textbf{NormED} is defined as
\begin{align*}
\text{NormED} = \frac{1}{N}\sum_{n=1}^{N} \frac{\text{EditDistance}(y^{(n)}, \hat{y}^{(n)})}{|y^{(n)}|},
\end{align*}
where \(y^{(n)}\) and \(\hat{y}^{(n)}\) here denote the ground truth and predicted \emph{transcriptions} of the $n$th test sample (character sequences, in contrast to their use as images elsewhere), \(\text{EditDistance}(\cdot)\) is the Levenshtein distance, \(|\cdot|\) denotes sequence length, and samples with an empty ground truth or prediction contribute zero. \textbf{CER} quantifies character-level recognition errors at the corpus level, $\text{CER} = \sum_{n} \text{EditDistance}(y^{(n)}, \hat{y}^{(n)}) \,/\, \sum_{n} |y^{(n)}|$. 

\begin{table}[h]
\begin{center}
\resizebox{0.48\textwidth}{!}{
\begin{tabular}{ccccc}
\toprule 
    \textbf{Dataset}    & \textbf{MNIST} & \multicolumn{3}{c}{\textbf{Saint Gall}} \\
\cmidrule(lr){1-1} \cmidrule(lr){2-2} \cmidrule(lr){3-5}
    \textbf{Method}     & \small \textbf{WAcc}(\%)~{$\uparrow$} & \small \textbf{WAcc}(\%)~{$\uparrow$} & \small \textbf{CER}(\%)~{$\downarrow$} & \small \textbf{NormED}~{$\downarrow$}\\
\midrule
    AHTR      & 99.34 (\(\pm\)0.05)  & 94.64 (\(\pm\)0.51)  & 1.26 (\(\pm\)0.12) &  0.0127 (\(\pm\)0.0013)\\
    G-AHTR    & 99.37 (\(\pm\)0.06)  & 94.40 (\(\pm\)0.37)  & 1.31 (\(\pm\)0.10)  &  0.0131 (\(\pm\)0.0009)\\
\bottomrule
\end{tabular}
}
\end{center}
\vspace{-0.1in}
\caption{Baseline recognition performance on the original MNIST and Saint Gall datasets, reported in WAcc(\%) for MNIST and WAcc(\%), CER(\%), NormED for Saint Gall. Results are averaged over 10 runs with standard deviation (\(\pm\)).}
\label{table_baseline}
\end{table}

\begin{table*}[t]
    \centering
    \resizebox{0.9\textwidth}{!}{
    \begin{tabular}{lccccccccc}
      \toprule
      Dataset 
      & \multicolumn{3}{c}{Synthetic MNIST} 
      & \multicolumn{3}{c}{Synthetic Saint Gall} 
      & \multicolumn{3}{c}{Synthetic Georgian} \\
      \cmidrule(lr){1-1} \cmidrule(lr){2-4} \cmidrule(lr){5-7} \cmidrule(lr){8-10}
      Methods & SSIM~{\small$\uparrow$} & PSNR~{\small$\uparrow$} & FID~{\small$\downarrow$} & SSIM~{\small$\uparrow$} & PSNR~{\small$\uparrow$} & FID~{\small$\downarrow$} & SSIM~{\small$\uparrow$} & PSNR~{\small$\uparrow$} & FID~{\small$\downarrow$} \\
      \midrule
      \cite{Starynska2021} & - & 17.781 & -
              & - & - & - 
              & - & - & - \\
      \midrule
      \cite{jampour2025revealing} & - & - & - 
              & - & - & - 
              & - & - & 144.68* \\
      \midrule
      Pix2pix & 0.9383 & \textbf{23.242} & 3.849 
              & 0.9855 & 27.771 & \textbf{11.066} 
              & \textbf{0.9913} & \textbf{37.619} & \textbf{20.367} \\
      \midrule
      cDDM1 & 0.8879 & 21.634 & 20.054 
            & 0.9804 & 26.238 & 21.285 
            & 0.9509 & 27.574 & 49.942 \\
      \midrule
      cDDM2 & \textbf{0.9419} & 23.238 & \textbf{3.837}
            & \textbf{0.9939} & \textbf{32.753} & 19.678
            & 0.9750 & 31.384 & 31.055 \\
      \bottomrule
    \end{tabular}
    }
    \vspace{-0.05in}
    \caption{Text separation results on synthetic palimpsest datasets using Pix2pix and cDDMs. Results from related works are included for comparison. *Note that \citet{jampour2025revealing} reported three document-level FID values, we therefore use their median in our table.}
    \label{tab:results_text_separation}
\end{table*}

\begin{table*}[!t]
    \centering
    \resizebox{0.9\textwidth}{!}{
    \begin{tabular}{ccccccc}
      \toprule
      Dataset & 
      & Synthetic MNIST 
      & \multicolumn{4}{c}{Synthetic Saint Gall}\\
      \cmidrule(lr){1-2} \cmidrule(lr){3-3} \cmidrule(lr){4-7}
      Methods & Scenario & WAcc~(\%)~{\small$\uparrow$} & WAcc~(\%)~{\small$\uparrow$} & WER~(\%)~{\small$\downarrow$} & CER~(\%)~{\small$\downarrow$} & Norm~ED~{\small$\downarrow$}\\
      \midrule
      \multicolumn{2}{c}{\cite{Starynska2021}} & 91.90 & - & - & - & -\\
      \midrule
      \multirow{3}{*}{AHTR} & Overlaid  & 97.71 (\(\pm\)0.12)
                                        & 79.63 (\(\pm\)1.48)
                                        & 20.37 (\(\pm\)1.48) & 5.75 (\(\pm\)0.43) & 0.0612 (\(\pm\)0.0054)\\
      \cmidrule(lr){2-7}
                            & Pix2pix   & 96.83 (\(\pm\)0.12)
                                        & 62.25 (\(\pm\)1.87)
                                        & 37.75 (\(\pm\)1.87) & 12.40 (\(\pm\)0.68) & 0.1430 (\(\pm\)0.0075)  \\
      \cmidrule(lr){2-7}
                            & cDDM2     & 97.83 (\(\pm\)0.11)
                                        & 91.70 (\(\pm\)0.73)
                                        & 8.30 (\(\pm\)0.73) & \textbf{1.95} (\(\pm\)0.25)& \textbf{0.0210} (\(\pm\)0.0030) \\
      \midrule
      \multirow{3}{*}{G-AHTR} & Overlaid & 97.68 (\(\pm\)0.15)
                                        & 78.30 (\(\pm\)1.38)
                                        & 21.70 (\(\pm\)1.38) & 6.18 (\(\pm\)0.48)& 0.0660 (\(\pm\)0.0055)\\
      \cmidrule(lr){2-7}
                              & Pix2pix & 96.81 (\(\pm\)0.11)
                                        & 61.58 (\(\pm\)0.86)
                                        & 38.42 (\(\pm\)0.86) & 12.59 (\(\pm\)0.50)& 0.1450 (\(\pm\)0.0070)\\
      \cmidrule(lr){2-7}
                              & cDDM2 & \textbf{97.97} (\(\pm\)0.08)
                                     & \textbf{91.78} (\(\pm\)0.83)
                                     & \textbf{8.22} (\(\pm\)0.83)& 1.98 (\(\pm\)0.24)& 0.0218 (\(\pm\)0.0027) \\
      \bottomrule
    \end{tabular}
    }
    \caption{Comparison of the average performance of AHTR and G-AHTR on Synthetic MNIST and Synthetic Saint Gall over 10 runs with standard deviation (\(\pm\)). Results are shown for direct recognition from the ``overlaid'' palimpsest images and for recognition after text separation using Pix2pix and cDDM2. cDDM1 is omitted due to its consistently lower separation performance compared to cDDM2.}
    \label{table_results_HTR}
\end{table*}

\subsection{Results}
\label{subsec:results}

Experimental results for text separation and recognition are reported across three scenarios highlighted in Section \ref{sec:exp}, as follows.

\paragraph{a. Baseline results:} Table \ref{table_baseline} presents the baseline recognition performance for the original MNIST and Saint Gall datasets. Since these datasets contain no overlapping texts, only recognition performance is reported. For each experiment, we performed 10 independent training runs and report the average model performance. 

The recognition of the Synthetic Georgian palimpsest is not considered here because its texts overlap, which makes a reliable baseline unavailable. Moreover, performing HTR on the Synthetic Georgian dataset is not feasible due to the lack of annotations, transcriptions, and training labels.

\paragraph{b. Direct recognition from overlaid palimpsest:} We then evaluate the performance of HTR methods in recognising the under-texts directly from the overlaid images in synthetic palimpsest datasets, following the exact same pipeline as we used in scenario (a). The results are reported in WAcc, CER, and NormED, listed in rows under scenario ``Overlaid'' in Table \ref{table_results_HTR}.

As can be observed in Table \ref{table_results_HTR}, performing HTR directly on synthetic palimpsests without first separating the text layers is, as expected, challenging due to the partial visibility of the under-text. However, evaluating the performance of our HTR models in this setting provides valuable insights and highlights the need for effective text separation, discussed below.

\begin{figure*}[!t]
    \centering
    \includegraphics[width=11.5cm]{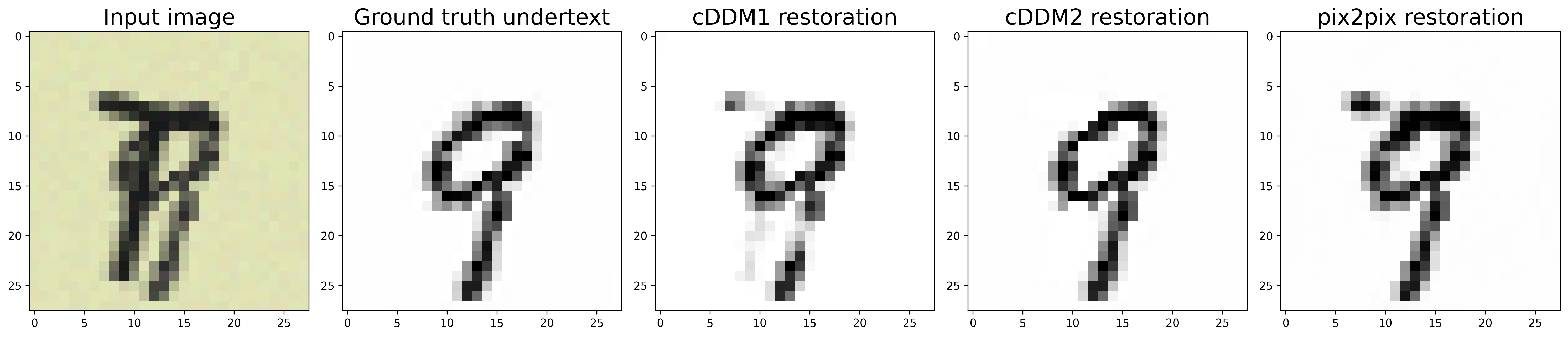}
    \includegraphics[width=11.5cm]{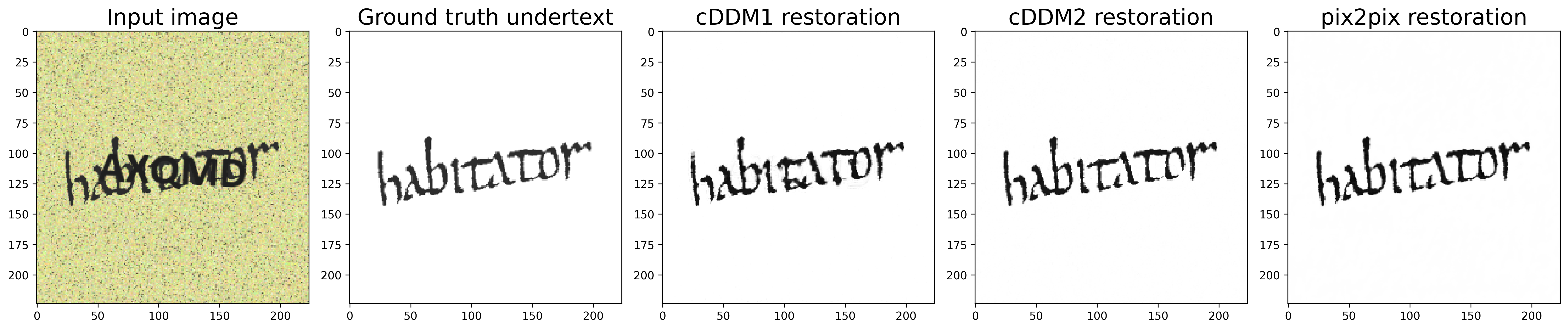}
    \includegraphics[width=11.5cm]{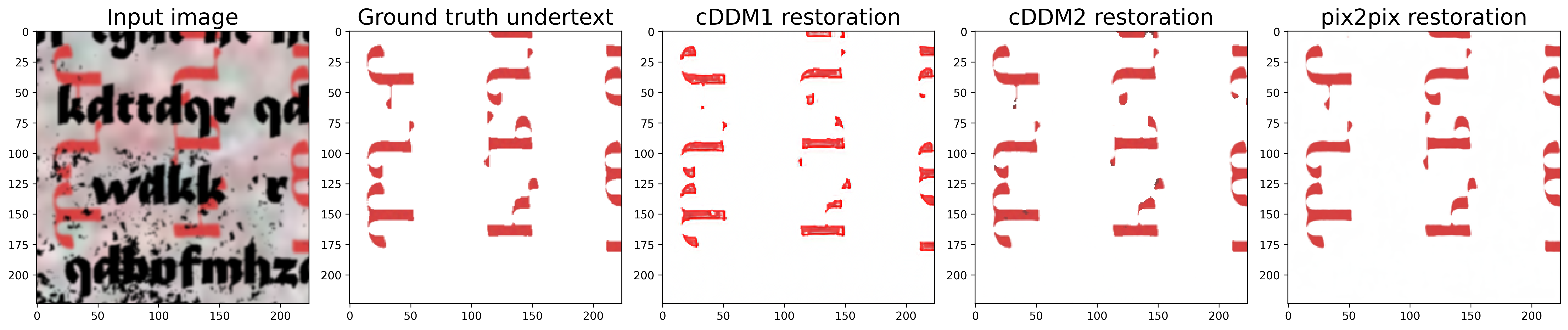}
    \caption{Examples of text separation results from different models, shown alongside the input palimpsest image and the ground truth under-text. The examples are drawn from the Synthetic MNIST, Synthetic Saint Gall, and Synthetic Georgian datasets (top to bottom).}
    \label{fig:text_sep_eg}
\end{figure*}

\paragraph{c. Recognition with text separation:} Furthermore, we integrate the text separation stage into the HTR pipeline, where Pix2pix and cDDMs are used to reconstruct the under-text from the palimpsest images. The models are trained on the respective training splits of each dataset, with hyperparameters (such as learning rate, epochs, and cDDM denoising steps) optimised to obtain the best models. The saved model checkpoints are then used to perform reconstructions on the test sets.

The separation results are evaluated using the three metrics introduced in Section \ref{subsec:metric}: SSIM, PSNR, and FID, as reported in Table \ref{tab:results_text_separation}. 

Overall, cDDM2 achieves the best performance on Synthetic MNIST and Synthetic Saint Gall, obtaining the highest SSIM on both datasets, together with the lowest FID and near-best PSNR on Synthetic MNIST and the highest PSNR on Synthetic Saint Gall (where Pix2pix retains the lowest FID), demonstrating the effectiveness of our proposed custom noise mechanism. Its performance on Synthetic Georgian is slightly lower than that of Pix2pix, primarily due to colour discrepancies. This can be observed in the examples in Figure \ref{fig:text_sep_eg}, where the cDDM reconstructions appear lighter than those of Pix2pix; however, the under-text shapes are clearly revealed and do not hinder recognition. 

The MNIST results (first row of Figure \ref{fig:text_sep_eg}) further reveal an interesting difference: Pix2pix often behaves more rigidly than cDDM by attempting to forcefully recover all visible details, such as the black dot above the true under-text ``9'', originating from the over-text ``T''. This rigidity contributes to Pix2pix’s consistently low FID. However, because FID assumes a Gaussian distribution--and, as discussed in Section \ref{subsubsec:cDDM} palimpsest noise is not always Gaussian, FID may be statistically biased \cite{chong2020effectivelyfid, jayasumana2024rethinkingfid}. Moreover, this tendency to reproduce irrelevant or misleading artifacts may negatively affect the downstream recognition stage.

The reconstructed under-text images are then used as inputs for HTR training and testing. Table \ref{table_results_HTR} reports the performance when text separation is applied prior to HTR. Since AHTR is designed for Latin characters only, its results are not compared with previous works involving other scripts, such as Greek, Syriac, and Hebrew \cite{madi2023scheme}. As a baseline observation, integrating under-text separation substantially improves recognition accuracy over direct HTR on raw palimpsests, particularly on the highly challenging Saint Gall dataset. Among the evaluated pipelines, our proposed cDDM2 + G-AHTR framework achieves the best overall performance, yielding the highest mean WAcc and lowest mean WER on both the Synthetic MNIST and Synthetic Saint Gall datasets, although the margins over cDDM2 + AHTR lie within one standard deviation of the ten runs. In particular, when paired with cDDM2 reconstructions, G-AHTR yields small improvements in word-level metrics (WAcc/WER), while CER does not improve (1.98 vs.\ 1.95 on Synthetic Saint Gall); with Overlaid and Pix2pix inputs, G-AHTR performs on par with or slightly below AHTR, suggesting that the guidance is most useful when the reconstructed under-text is of high quality. This disparity is expected, as the high variability and ambiguity inherent in handwritten text make character-level recognition more challenging. It is worth noticing that the language guiding mechanism comes at no cost on training stability, since the standard deviations of G-AHTR results are around the same level as those of AHTR.

\begin{table*}[ht!]
\centering
\begin{tabular}{lllllll}
\toprule

AHTR & 
\includegraphics[width=.12\linewidth,valign=m]{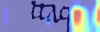} & 
\includegraphics[width=.12\linewidth,valign=m]{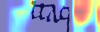} & 
\includegraphics[width=.12\linewidth,valign=m]{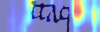} & 
\includegraphics[width=.12\linewidth,valign=m]{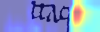} & 
\includegraphics[width=.12\linewidth,valign=m]{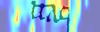} & 
\includegraphics[width=.12\linewidth,valign=m]{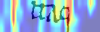}\\ 

\midrule

G-AHTR & 
\includegraphics[width=.12\linewidth,valign=m]{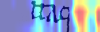} & 
\includegraphics[width=.12\linewidth,valign=m]{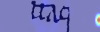} & 
\includegraphics[width=.12\linewidth,valign=m]{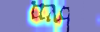} & 
\includegraphics[width=.12\linewidth,valign=m]{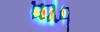} & 
\includegraphics[width=.12\linewidth,valign=m]{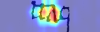} & 
\includegraphics[width=.12\linewidth,valign=m]{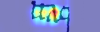} \\

\bottomrule
\end{tabular}
\caption{Comparison of activation maps on example under-text ``itaq'' obtained from AHTR and G-AHTR during training, on every third epoch (i.e. epochs 1, 4, 7, 10, 13 and 16). The activation maps from G-AHTR show clear signs of convergence toward the CLIP map.}
\label{tab:train_maps}
\end{table*}

While the absolute performance gains provided by the language-guided supervision are modest, we hypothesise that the performance ceiling is constrained by the following primary factors: i) the inherent limitations of CLIP in fine-grained handwriting recognition, which become more obvious when the intermediate recovered under-text remains highly degraded; ii) gradient competition between the primary CE loss and the auxiliary activation constraint, as strict spatial alignment with CLIP's feature maps can be partially orthogonal to the optimal feature representations needed for pure character decoding.

\begin{figure}[!t]
    \centering
    \begin{subfigure}[b]{0.25\textwidth}
        \centering
        \includegraphics[width=1.2in]{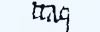}
        \captionsetup{font=small}
        \caption{cDDM}
    \end{subfigure}%
    ~ 
    \begin{subfigure}[b]{0.25\textwidth}
        \centering
        \includegraphics[width=1.2in]{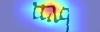}
        \captionsetup{font=small}
        \caption{CLIP}
    \end{subfigure}
    \caption{Under-text image from cDDM and the activation map from CLIP for ``itaq''.}
    \label{fig:act_maps}
\end{figure}

To further illustrate the contribution of the vision-language (CLIP) component, Figure \ref{fig:act_maps} visualises the ground truth activation map from the fine-tuned CLIP model, and Table \ref{tab:train_maps} shows the activation maps computed from AHTR and G-AHTR during training on an arbitrarily selected example. These maps show that the CLIP-guided activation constraint encourages G-AHTR to focus more strongly on the under-text regions while suppressing irrelevant artifacts and noise. This shift in activation toward semantically meaningful regions provides more interpretable model behaviour and, in the cDDM2 setting, coincides with small word-level accuracy gains. In particular, in challenging cases where AHTR hallucinates the word endings, G-AHTR tends to make more reasonable predictions (see Table \ref{examples}). 

These findings deepen the understanding of how automatic methods could be applied and interpreted when developing and evaluating palimpsest reconstructions. Our G-AHTR method introduces a new paradigm in cross-modal regularisation by anchoring the HTR model's visual attention to CLIP's vision-language alignments via GALS (Guiding visual Attention with Language Specification) \cite{petryk2022guiding}. This framework offers a foundational guidance mechanism that can be extended across diverse domains.

\section{Conclusion}
\label{sec:concl}
This work advances automated palimpsest recovery by integrating structured noise modelling, generative reconstruction, and language-grounded recognition. We design a custom denoising diffusion model to capture layered degradation, explore Pix2pix GANs for restoration, and incorporate textual priors from vision-language attention maps for under-text recognition. Experiments demonstrate improved performance across diverse benchmarks. In particular, the cDDM2 + G-AHTR pipeline achieves the best word-level results and competitive overall scores, while yielding more interpretable attention behaviour. Performance variations across the synthetic MNIST, Saint Gall, and Georgian datasets stem from differences in ink properties, script complexity, and background noise, influencing the suitability of the GAN, diffusion, and HTR components. To support future research and standardised benchmarking, we release synthetic datasets, CLIP-compatible annotations, and pre-trained models.

\begin{table}[!t]
\begin{center}
\resizebox{0.49\textwidth}{!}{
\begin{tabular}{lll}
\toprule 
       Original label & AHTR & G-AHTR \\
\midrule
    impossibile   & impossibilitie & impossibile  \\
    solitudinis    & solitudiiis   & solitudinis  \\
    solatium    & solattuum  & solatium  \\
    construens    & construeninininininininin   & construeni  \\
    linea    & lineama   & linea \\
\bottomrule
\end{tabular}
}
\end{center}
\vspace{-0.1in}
\caption{Example predictions from Synthetic Saint Gall (in Latin) show G-AHTR is less likely to hallucinate than AHTR, highlighting the advantage of language-guided HTR.}
\label{examples}
\end{table}

Future work includes exploring other HTR architectures and generalisation, though we used AHTR for its ease of use and reproducible code. We plan to extend evaluations from word-level to line- and page-level, test on real palimpsests such as the Archimedes dataset, and fine-tune the CLIP model on large-scale handwriting corpora. Finally, we aim to establish a comprehensive benchmark for palimpsest research, including curated datasets, methods, and evaluation metrics.


\section{Acknowledgments}
This research was supported by the Kjell and Märta Beijer Foundation, and the Wallenberg AI, Autonomous Systems and Software Program (WASP) funded by the Knut and Alice Wallenberg Foundation. Computational resources were provided on the Berzelius system funded by the Knut and Alice Wallenberg foundation and operated by NAISS. The authors thank Prashant Singh and Anna Starynska for valuable discussions and feedback, and the Action Editor and reviewers for their constructive comments.



\bibliography{tacl2021}
\bibliographystyle{acl_natbib}

\iftaclpubformat

\onecolumn

%
%
%
%
%
%
  

\end{document}